\documentclass[10pt,conference,a4paper]{IEEEtran}
\ifCLASSINFOpdf
  \usepackage[pdftex]{graphicx}
\else
\fi
\usepackage{multirow}

\usepackage{amsmath}
\DeclareMathOperator*{\argmax}{arg\,max}
\usepackage{algorithm2e}
\newcommand\blfootnote[1]{%
  \begingroup
  \renewcommand\thefootnote{}\footnote{#1}%
  \addtocounter{footnote}{-1}%
  \endgroup
}

\usepackage{xcolor}
\usepackage{comment}

\newcommand*{\lbr}{\left(}
\newcommand*{\flbr}{\!\lbr}
\newcommand*{\rbr}{\right)}

\begin{document}
%
\title{Exploiting the Logits: Joint Sign Language Recognition and Spell-Correction}

\author{%
\IEEEauthorblockN{Christina Runkel$^*$, Stefan Dorenkamp$^*$, Hartmut Bauermeister, Michael Moeller}
\IEEEauthorblockA{Department for Computer Science and Electrical Engineering, University of Siegen \\
Emails: \{christina.runkel, stefan.dorenkamp\}@student.uni-siegen.de} \{hartmut.bauermeister, michael.moeller\}@uni-siegen.de\\
}

%


\maketitle

\blfootnote{$^*$equal contribution}
\begin{abstract}
Machine learning techniques have excelled in the automatic semantic analysis of images, reaching human-level performances on challenging benchmarks. Yet, the semantic analysis of videos remains challenging due to the significantly higher dimensionality of the input data, respectively, the significantly higher need for annotated training examples. By studying the automatic recognition of German sign language videos, we demonstrate that on the relatively scarce training data of $2.800$ videos, modern deep learning architectures for video analysis (such as ResNeXt) along with transfer learning on large gesture recognition tasks, can achieve about $75\%$ character accuracy. Considering that this leaves us with a probability of under $25\%$ that a $5$ letter word is spelled correctly, spell-correction systems are crucial for producing readable outputs. The contribution of this paper is to propose a convolutional neural network for spell-correction that expects the softmax outputs of the character recognition network (instead of a misspelled word) as an input. We demonstrate that purely learning on softmax inputs in combination with scarce training data yields overfitting as the network learns the inputs by heart. In contrast, training the network on several variants of the logits of the classification output i.e. scaling by a constant factor, adding of random noise, mixing of softmax and hardmax inputs or purely training on hardmax inputs, leads to better generalization while benefitting from the significant information hidden in these outputs (that have  $98\%$ top-5 accuracy), yielding a readable text despite the comparably low character accuracy.  
\end{abstract}


%
\IEEEpeerreviewmaketitle

\section{Introduction}
The automatic recognition and translation of sign language with handheld cameras on mobile devices bares a great potential to impact our social life, as it would enable the seamless communication with dumb people. While deep learning techniques have revolutionized the field of computer vision over the last decade, significant challenges remain in the automatic analysis of video data due to their high dimensionality as well as the comparably scare training data available to train activity recognition systems on specific tasks such as sign language understanding. 

In this paper we consider the easier problem of classifying videos of the German sign language alphabet with videos from the RWTH Fingerspelling Database \cite{Imran2019}, which are recorded on a tripod. Despite advances in network architectures (e.g. the ResNext approach for video analysis as considered in \cite{realtimeHandGesture}), data augmentation, and despite actively exploiting transfer learning approaches by pretraining on the Jester V1 dataset consisting of 148.092 videos of 27 different activities, our approach achieves a character accuracy of $75\%$ only. While this is impressive in comparison to video analysis systems from 10 years ago, and comparable to the most recent works that tailored and optimized their networks on this specific task, such accuracies are insufficient to produce readable text: Assuming a uniform random distribution of accuracies, a $5$ letter word is spelled correctly with a probability of only $(0.75)^5 \approx 0.237\%$. This is the reason why spell-correction systems that additionally learn a specific language are and will remain crucial in such applications. 

\begin{figure}[tb]
	\centering\includegraphics[width=1\linewidth, trim={1cm 1cm 0 1cm},clip]{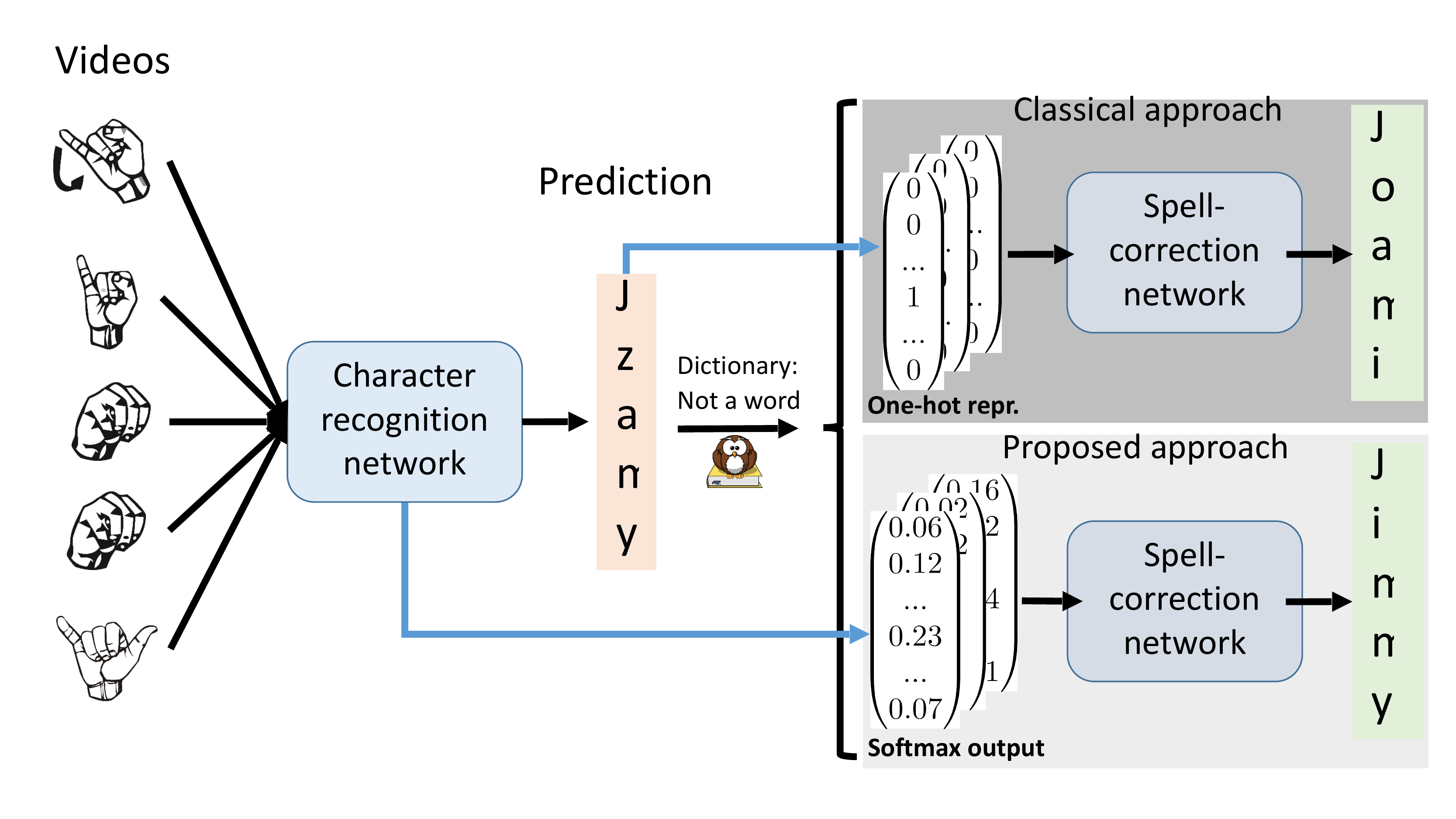}
	\caption{Classical and proposed spell-correction approach: While a typical spell-correction method just depends on the input word and not on the method/user that generated the word, we propose to exploit the softmax output of the character-prediction network and demonstrate that significantly higher word accuracies are possible with such an approach. }
	\label{fig:teaser}
\end{figure}

In this paper we show that despite the rather low accuracy of the final character recognition, the classification network almost always manages to narrow down the number of possible characters from $35$ ($26$ characters plus $3$ German \textit{Umlaute}, the letter sequence SCH and the numbers from 1 to 5) to about $5$ (more ambiguous) possible signs, as our network reaches a top-5 accuracy of $98\%$. This, however, means that the classification network provides significantly more information than just a (possibly misspelled) word. Therefore, we propose to use the network's softmax output (rather than a binary classification) as an input to a spell-correction network. By using softmax instead of hardmax vectors as an input to our spell-correction network, and combining it with a dictionary look-up as well as a traditional spell-correction approach, we are able to improve the overall system's word accuracy significantly, e.g. from $22\%$ when using a pure video recognition network to $75\%$ with our entire system. Figure \ref{fig:teaser} illustrates the core idea of the softmax-based spell-correction.

\section{Related Work}
Activity and gesture recognition -- as an enabler for a wide range of technologies -- is a well-studied field of research in computer vision. In addition to the classification based on video data or image sequences, for which \cite{asadi2017survey} provides an overview, many approaches work with additional (body worn) sensors \cite{wang2019deep} or other tools, which sometimes include special cameras \cite{yang2014super}. While gesture recognition in general concerns the recognition of arbitrary previously defined gestures and thus is a wide field (including applications such as the control of technical devices or for cooperation between humans and machines), we will now focus on the classification of sign language and fingerspelling data only.

\subsection{Fingerspelling}
There are already various approaches in the field of sign languages, which can be classified into five groups. The first group includes those that work with tools such as a painted glove \cite{Dong2015}, a Leap Motion Controller \cite{Mohandes2014, Tao2018} or special recordings such as depth images \cite{Neumann2018}. The second approach uses simple RGB images, i.e. single frames, which usually do not allow for the recognition of moving gestures (e.g. \cite{Kasukurthi2019, Bheda2017}). In particular, for the German finger alphabet we consider in this paper, some characters (like 'i' and 'j' or 'a' and '\"a') are indistinguishable on single frames. In the third group of more classical approaches fuzzy logic and hidden Markov models, e.g. \cite{Starner1998, Kishore2012}, are used to identify characters. The fourth group performs sign recognition using video, where a sign already represents a word \cite{Cui2017, Huang2018, Jing2019}, which yields significantly more classes and thus demands even more training data. 
In the fifth and last group the classification is performed on videos of the fingerspelling alphabet, where neither tools nor special cameras are necessary, see e.g. \cite{Papadimitriou2019, Shi2018}. The state-of-the art for such approaches is to exploit the expressiveness of deep convolutional neural networks that act on the 3D spatio-temporal volume of the input videos. Some works additionally extract prior information such as optical flow fields, e.g. in \cite{shi2019fingerspelling}. The faithful recognition of the fingerspelling alphabet becomes particularly challenging in the case of rather scarce training data, e.g. when working with the RWTH Fingerspelling Database \cite{dreuw06smvp}, in which $80$ videos are available for each character from two different perspectives. To our best knowledge, the highest recorded accuracy on the aforementioned data set was achieved by \cite{Imran2019} when limiting the approach to one fixed perspective, and amounts to $85\%$.

\subsection{Spell-Correction}
Spell-correction can be thought of as finding the most probable word $c \in C$ for a given misspelled word $w$ among all possible correctly spelled words $C$ that exist in the considered language. According to Bayes rule this task can be seen as maximum a posteriori problem in the form of
\begin{equation}
\label{eq:bayeslaw}
    \argmax_{c} P \flbr c \mid w \rbr = \argmax_{c} P \lbr w \mid c \rbr P \flbr c \rbr,
\end{equation}
i.e., the probability $P\lbr c \mid w \rbr$ that $c$ is the correct word under the assumption that we observed $w$ can be expressed as the product between the probability $P(c)$ that $c$ is a used word and the probability $P\lbr w \mid c \rbr$ that $w$ is the misspelled word under the condition that $c$ is the intended word. 

Recent works in the field of spell-correction can be classified mainly into two groups - statistical approaches and approaches which make use of Deep Learning techniques. 

Statistical spell-correction approaches exploit \eqref{eq:bayeslaw} and explicitly model probabilities $\tilde P(c)$ based on word frequencies and $\tilde P \lbr w \mid c \rbr$ based on a misspelling process. The latter conditional probabilities mainly rely on editing distances, e.g. \cite{Gezmu2018, Gupta2019, Beeksma2018}, and n-grams, e.g. \cite{MashodRana2018, Dashti2018}.

In contrast, Deep Learning techniques attempt to directly learn a corrected word $c$ by approximating the mapping $w \mapsto \argmax_{c} P \flbr c \mid w \rbr$ by a feed-forward network $N$ with parameters $\theta$ such that
\begin{equation}
	\label{eq:nn}
	N(w;\theta) \approx \argmax_{c} P \flbr c \mid w \rbr.
\end{equation}
For spell-correction tasks, mainly Recurrent Neural Networks \cite{Sakaguchi2016, Li2018} with a Decoder-Encoder architecture \cite{Etoori2018, Xie2016, Zhou2019} are used. As the tasks of Machine Translation and Grammatical Error Correction are partly similar to spell-correction, approaches like \cite{Zhou2019, Grundkiewicz2018} make use of common techniques in these areas, which led to the use of convolutional neural networks (CNNs) for spell-correction, e.g. in \cite{Chollampatt2018, Kim2016}, to convincingly model short-term dependencies. 

In all the above approaches, the word $w$ is represented by a sequence of letters, each of which is encoded in a one-hot representation. In other words, each character becomes a unit vector whose length is equal to the overall number of characters in the alphabet. A vector that has a $1$ in the $i$-th entry and $0$ in all other entries represents the $i$-th character of the alphabet.


\section{Proposed Approach}
\subsection{Character Gesture Recognition and Initial Word Prediction}
In this work we consider the translation of a fingerspelling video into written text using convolutional neural networks. Our main goal is to highlight the advantages of exploiting the softmax outputs of the video classification network to obtain improved spell-correction results. Thus, we do not consider the problem of dividing a video stream into different characters, but rather assume that such a division as well as the information which character videos form a word is provided. We furthermore assume the input word to be error-free i.e. the sequence of videos of spelled characters forms a correctly spelled German word. For the character recognition network we use a the ResNeXt-101 implementation of \cite{realtimeHandGesture} pretrained with the Jester V1 data set \cite{20bnJester} that consist of 148.092 videos of $27$ different hand gestures. None of the hand gestures, however, represents a letter of the finger alphabet.

\begin{figure}[ht]
	\centering\includegraphics[width=0.5\linewidth]{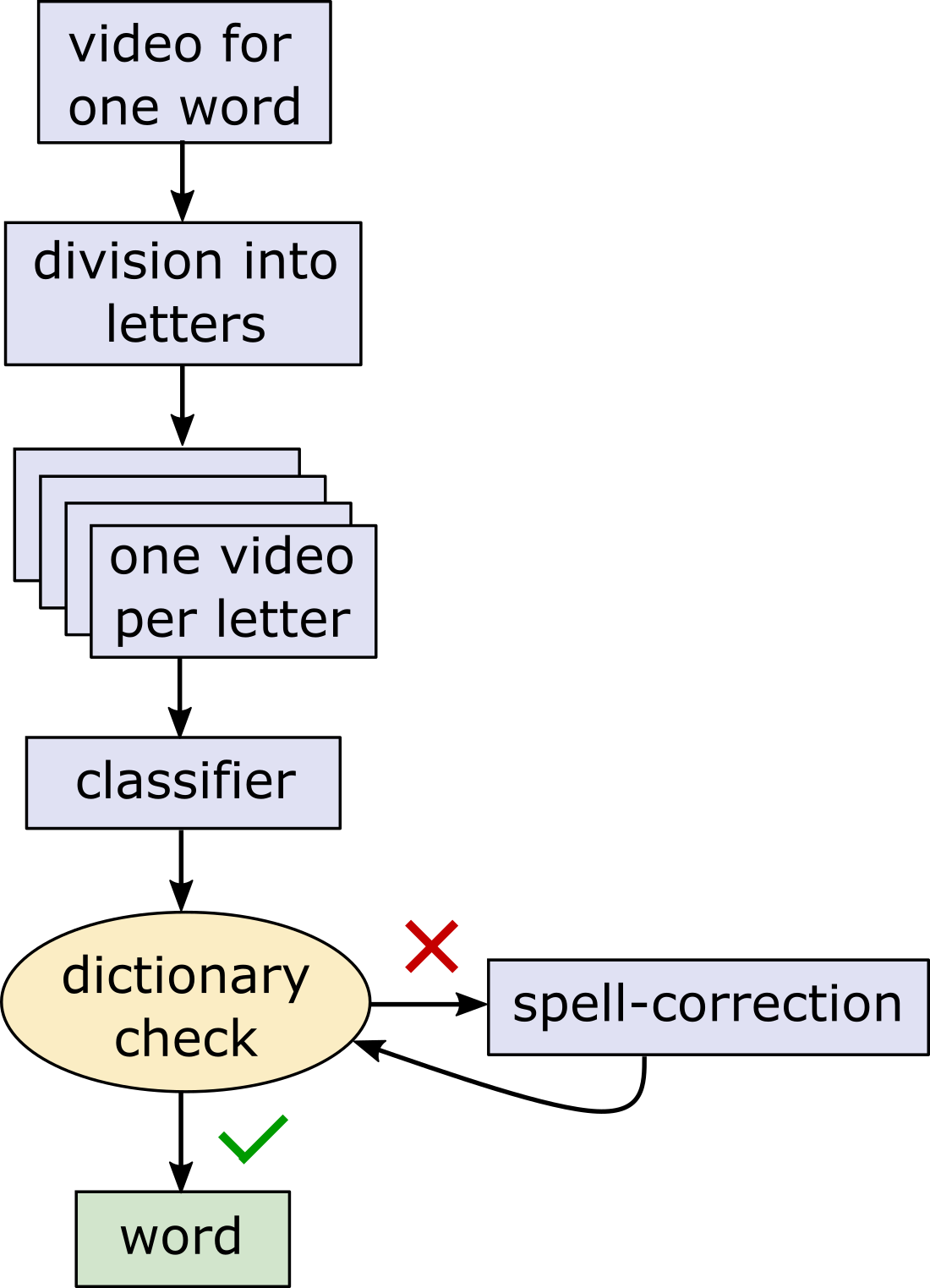}
	\caption{Workflow of the proposed approach from video input to word output. A video for a word is divided into separate letters which are classified separately, and reassembled to a word. Subsequently, the softmax probabilities of the classifier go into our spell-correction approach, if the word cannot be found in a dictionary.}
	\label{fig:pipeline}
\end{figure}

As illustrated in Figure \ref{fig:pipeline}, during testing we feed all character videos that assemble a word into the trained classification network and assemble a predicted word. Subsequently we use a dictionary to predict whether such a word exists in the German language. More specifically, we make use of the Free German Dictionary \cite{Schreiber2019} which has originally been used for the Open Source spell-correction GNU Aspell. It consists of more than 1.9 million entries for the German language, which are sorted alphabetically. If the predicted word can be found in the dictionary, it is assumed to be correct. Otherwise, it is fed into the spell-correction system to be detailed in the next subsection. 


\subsection{Spell-Correction}
\label{sec:spellCorrection}
\begin{figure}[ht]
	\centering\includegraphics[width=\linewidth]{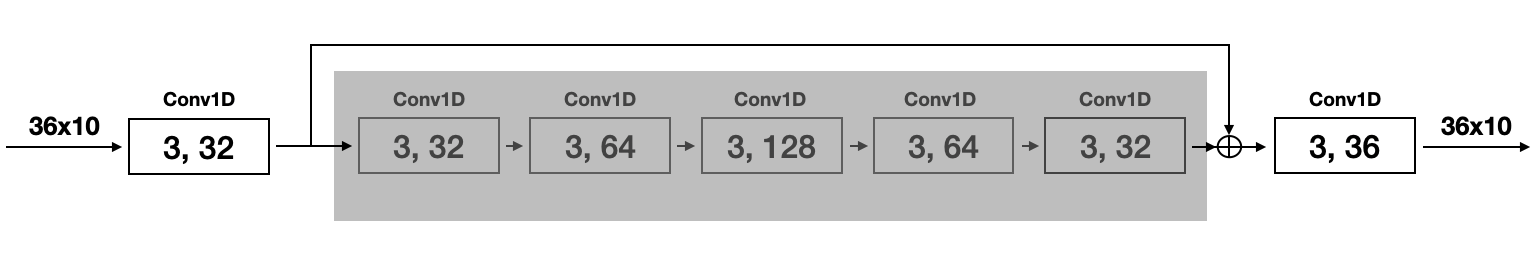}
	\caption{Architecture of the spell-correction network. Input and output are $36 \times 10$ matrices. For every convolutional layer, $(k, c)$ denotes the kernel size $k$ and the number of output channels $c$. After each layer, a LeakyReLU activation function and batch normalization \cite{batchnorm} are used.}
	\label{fig:conv1net_architecture}
\end{figure}

We consider two algorithmic approaches in our spell-checking experiments:
\begin{enumerate}
    \item \textit{Neural spell-correction:} In case the dictionary lookup fails, we propose to first exploit a machine learning based spell-correction technique, with Figure \ref{fig:conv1net_architecture} illustrating the architecture of the spell-correction network we used. As the input of the network is assumed to contain mostly correctly classified letters, a ResNet-like architecture with skip connections is utilized to improve the flow of the gradient and simplify learning the identity. In addition to the convolutional layers, the network uses LeakyReLU activation functions and batch normalization after each layer. As an average German word has word length 6, we fixed the input size to a maximal word length of 10 letters.
    \item \textit{Statistical spell-correction (optional):} As we will detail in the numerical results, the output of the CNN-based spell-correction is often more readable than a competing statistical spell-correction, but did not necessarily yield words from a dictionary. On the contrary, the statistical spell-correction failed for heavily erroneous words, but gave correctly spelled ones for words with minor (single character) mistakes, which motivated the use of a statistical spell-correction on the output of the spell-correction network. 
    \begin{algorithm}[h]
	\label{algorithm:norvig}
	\caption{Spell-correction according to Norvig \cite{Norvig2016} using Levenshtein distance $D$.}
	\uIf{$w\in C$} {\Return $w$}
	\uElseIf{there $\exists c_1\in C$  s.t. $D_{w,c_1} = 1 $}{\Return $\arg \max_{D_{w,c_1} = 1 ~ p(c_1)}$ }
	\uElseIf{there $\exists c_2\in C$  s.t. $D_{w,c_1} = 2 $}{\Return$\arg \max_{D_{w,c_2} = 2~ p(c_2)}$}
	\Else{\Return $w$}
    \end{algorithm}
    The statistical spell-correction used is the algorithm of Norvig \cite{Norvig2016} (see Algorithm \ref{algorithm:norvig}). It computes the most likely correction $c$ in a candidate set $C$ of words of a language by maximizing the probability $p(c|w)$ of a correction candidate $c$ when given a misspelled word $w \in W$ according to Bayes theorem \eqref{eq:bayeslaw}.
    To compute the conditional probability $p(w|c)$, the algorithm makes use of the Levenshtein distance \cite{Levenshtein1965}, which measures distances by the number of simple letter manipulations required to get from one word to another. Simple letter manipulations include replacement, insertion and deletion of letters of a word. Algorithm \ref{algorithm:norvig} shows the algorithm of Norvig for a maximal Levenshtein distance of two. A first step checks, whether the ``misspelled'' word $w$ is already in the set of correction candidates $C$. If it is not in the set of correction candidates, the algorithm checks for existence of a correction candidate $c_1 \in C$, such that the Levenshtein distance $D_{w,c_1}$ of the misspelled word $w$ and the correction candidate $c_1$ equals 1. This part of the algorithm can be repeated arbitrary often, with increasing Levenshtein distance, to find any possible correction candidate. However, as searching for a correction candidate with a high Levenshtein distance becomes more and more computationally intensive, a commonly used maximal Levenshtein distance is $D_{w,c}=2$. If no correction candidate can be determined, the algorithm returns the misspelled word $w$.
\end{enumerate}
\section{Generalization strategy} \label{section:gen_strategy}

\subsection{Fingerspelling network}
During our numerical experiments we found the small amount of training data to be the limiting factor for obtaining more accurate classification results: Even significantly simpler 3D CNNs quickly resulted in good training accuracy, but did not generalize.

Therefore, we exploited transfer learning on the significantly larger Jester V1 data set, that is reasonably similar to the desired task, as it aims to classify different hand gestures. In addition, we augmented the training data by exploiting random rescaling by a factor out of $\left \{ 1, \frac{1}{2^{0.25}}, \frac{1}{2^{0.5}}, \frac{1}{2^{0.75}}, \frac{1}{2} \right \}$ along with cropping back to 112x112 pixels, randomly at one of the four corners or in the middle. We found such an augmentation to reduce overfitting and encourage an invariance with respect to scale and position of the actors in the scene. 


\subsection{Spell-correction network}
As the spell-correction network uses the output of the classification network as input, its training data is equally limited. While on the one hand, the main point of this paper is to exploit the impressive top-5 accuracy of up to $98\%$ of the classification network by feeding the softmax-output into the spell-correction network, we found that the limited amount of training data quickly makes the spell-correction network learn by heart which softmax output corresponds to a certain character. The alternative of using hardmax outputs reduces to the classical one-hot representation of characters we discussed in the related work, but of course allows to generate arbitrary amounts of training data. We therefore conduct an ablation study with different types of representations of the letters, including a \textit{hardmax} representation, i.e., representing each character as a one-hot vector, a \textit{softmax} representation, i.e., using the output of the classification network for a real character video as an input to the spell-correction network, and four different variants to mix or augment these representations. 

While the above representations refer to the input to the spell-correction network \textit{during training}, in our numerical experiments we additionally experiment with using hardmax or softmax representations \textit{during testing}. 

\section{Implementation}
\subsection{Data sets}
We use two video data sets to train and evaluate our approach. One is the well-established RWTH German Fingerspelling Database, and one is a self-recorded data set in the spirit of the first to test how well the approach generalizes to different actors/people who do the fingerspelling as well as to a different recording location. 

\subsubsection{RWTH German Fingerspelling Database}
The German Fingerspelling Database of the RWTH Aachen University is freely available and contains a total of 3000 videos on all 35 gestures of the German finger alphabet. The recordings have a resolution of 320x240 pixels or 352x288. Half of the videos show only the hand (hereinafter referred to as \textit{R-Cam1}), while the other half also includes the person (hereinafter referred to as \textit{R-Cam2}). Furthermore, twenty people were involved. We refer to \cite{dreuw06smvp} for details.

\subsubsection{Self-recorded data}
To investigate the generalization of the proposed framework, a separate test data set with 420 videos was created. The recordings were made in two different positions. Similar to the RWTH dataset, it consists of videos from two different perspectives. In the first position only the hand and arm are shown, and in the other position additionally the upper body and the face are visible. In each position, three people recorded every character of the fingerspelling alphabet twice. 

\subsection{Character classification network}
The character classification network is implemented in PyTorch. After loading the weights from the training with the Jester V1 data set, the last fully connected layer is replaced by a new one that has 35 output neurons. During the further training all layers remain trainable and are not frozen.

We'd like to point out that the network is trained directly on the ground truth letter corresponding to the input videos, i.e.\ it is not trained jointly with the spell-correction network, because such an approach could lead to intentional wrong (or even uninterpretable) predictions of the classification network that are learned to be correct by the spell-checker. 

 As an optimizer we use stochastic gradient descent initialized with a momentum of 0.9, a dampening factor of 0.9 for momentum and a weight decay of 0.001. We use an initial learning rate of 0.01, and reduce it by a factor of 10 in the 10th and 25th epochs. 
 
 As a further pre-processing step, the input video is first converted into 32 frames. For this purpose, the total number of frames is divided by 32 and rounded off to calculate a dynamic step size $s$. Subsequently, we use the first frame and go through the frames with the step size $s$. If there is one frame too few, the last frame is used twice. We verified visually that the individual gestures are well reproduced and visible by this conversion. Finally, we normalize the input data to have zero mean and unit variance.

Before the training starts, the entire data set is shuffled randomly. From the total data set, 80\% is used for training and 20\% for validation. The validation is done after each epoch and the total of epochs we trained is 25.

\subsection{Spell-Correction Network}
\begin{figure}[ht]
	\centering\includegraphics[width=0.8\linewidth]{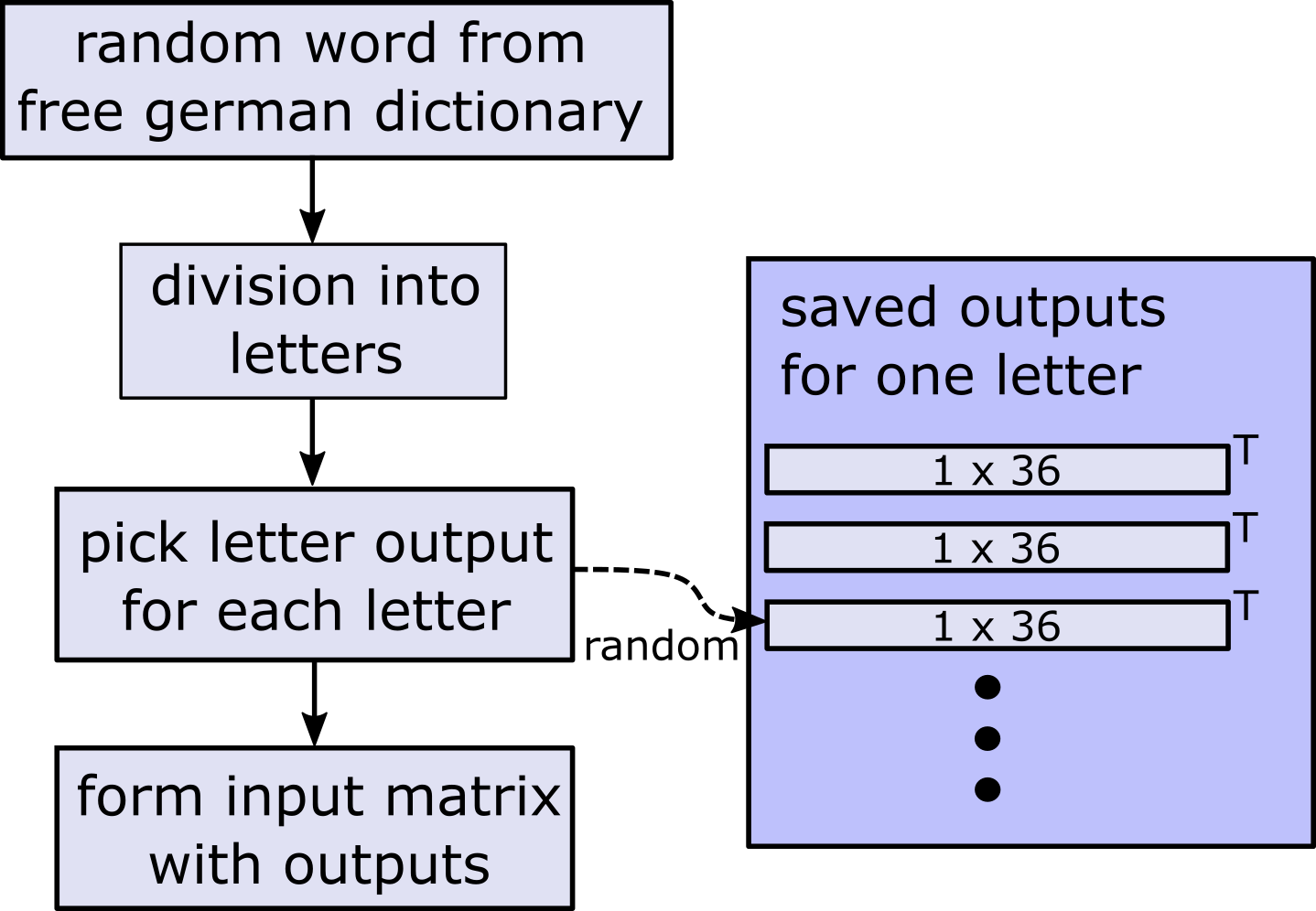}
	\caption{Illustrating the training data formation process for the spell checking network: A word is divided into separate letters for each of which we select a random output of the classification network of exactly this letter. The collection of all letters of the word result in a matrix which form the input to the spell-checking network. }
	\label{fig:spelling_correction_process}
\end{figure}

The spell-correction network is implemented in TensorFlow, and the data generation for its training process is illustrated in figure \ref{fig:spelling_correction_process}. As a first step, a word is picked randomly from the Free German Dictionary \cite{Schreiber2019}. After separating the word into letters, we randomly select a (buffered) output of the classification network for an input video that corresponds to the correct letter. The data used to produce such outputs coincides with test data used for the classification network. Finally, for a word of length $n$, the $36 \times 1$ vectors of the separate letters are rearranged into a $36 \times n$ matrix, which serves as an input to our spell-correction CNN described in section \ref{sec:spellCorrection}. Note that while our classification network outputs vectors of length $35$, we append an additional zero-entry for an `out of vocabulary (OOV)'' class, which can be used to identify uncertainty of individual letters. In particular, we use it to represent missing characters, such that we can pad the input word to have a fixed length of 10.

As explained in section \ref{section:gen_strategy}, we perform an ablation study with the spell-correction inputs coming from a hardmax-, softmax-, or mixed hardmax and softmax of the classification network.
Independently of the type of input, the dataset consists of 9830 training pairs i.e. pairs of misspelled and ground truth words. It is trained for 100 to 250 epochs, depending on the type of input data (hardmax-, softmax-, mixed-inputs) with a batch size of 1024, a learning rate of 0.001 and Adam \cite{adam} as an optimizer.

\subsection{Statistical Spell-Correction}
For statistical spell-correction, the spell-correction algorithm of Norvig, implemented within the Python library ``pyspellchecker'' \cite{Barrus2019}, is used. It comes with a default word frequency dictionary for the German language and variable Levenshtein distance. For the present use case, a Levenshtein distance of two has proven to be most suitable.

\section{Numerical experiments}
For our numerical evaluation we define four different test cases that ought to demonstrate the systematic improvements of the proposed approach:
\begin{itemize}
    \item \textbf{Test Case 1 (TC1)} uses the first half of the RWTH data set of perspective 1 for training the classification network, uses the same data to generate outputs of the (trained) classifier which are used for training the spell-correction, and finally tests on the second half of the same perspective. 
    \item \textbf{Test Case 2 (TC2)} takes the same approach as TC1, but jointly on both perspectives, i.e., it takes the first half of the entire RWTH data set for training the classifier, uses the same data to generate training samples for the spell-correction, and finally tests on the second half of the RWTH data set, always using both perspectives. 
    \item \textbf{Test case 3 (TC3)} uses all videos of the RWTH data set that are taken from the first perspective as well as one video from each actor, each character and each perspective as an input. The training data for the spell-checking network is created by feeding augmented versions of the of our own training videos into the classifier, and finally the framework is tested with all remaining (unseen) self-recorded videos. 
    \item \textbf{Test case 4 (TC4)} is identical to TC3 except that the entire RWTH data set is used for training, such that significantly more data (with different perspectives) is available. 
\end{itemize}
We chose the above settings to study the behavior of fixed viewpoints vs. variable viewpoints (reflected by TC1 and TC3 vs. TC2 and TC4), and study reusing the training data of the classifier to generate training data for the spell-checker vs. generating new outputs via data augmentation (reflected by TC1 and TC2 vs. TC3 and TC4),.

\subsection{Letter Classification}
Tabular \ref{tabular:evaluation_fingerspelling_words} shows the classification accuracy achieved by the plain classification network (including data augmentation and transfer learning as described in Section \ref{section:gen_strategy}). As we can see, among the methods TC1 and TC2 that considered the RWTH data only, training on a specific viewpoint/perspective seemed significantly easier than handling variable perspectives. This situation, however, changes when considering our self-recorded data in TC3 and TC4, where the variable perspective in TC4 gives higher accuracy than the fixed one of TC3. We conjecture that the self-recorded data had less variability between the perspectives which allowed the network to benefit from the additional training data. This effect seemed to have outweigh the challenge of overcoming the change of perspective. 
\bgroup
\def\arraystretch{1.5}
\begin{center}
\begin{table}[!ht]
	\centering
	\begin{tabular}{|l|c|c|}
	\hline
	\textbf{Test case} & \textbf{Character-Accuracy} & \textbf{Word-Accuracy}  \\ \hline \hline
	TC1 & 74~\% & 23~\% \\\hline
	TC2  & 58~\% & 14~\% \\\hline
	 TC3 & 67~\% & 18~\% \\\hline
	 TC4 & 73~\% & 22~\% \\\hline
	\end{tabular} \newline
	\caption{Evaluation of character and word accuracy of the classification network.}
	\label{tabular:evaluation_fingerspelling_words}
\end{table}
\end{center}
\egroup

\subsection{Text Prediction}
The full workflow of our joint sign language recognition and spell-correction approach (see figure \ref{fig:pipeline}) is evaluated by testing the accuracy of 100 randomly generated German words, by assembling random videos of the corresponding characters from the set of test videos of our four different scenarios TC1 through TC4. We evaluate six different versions of our spell-correction network:
\begin{itemize}
    \item \textbf{H/H} trains the spell-correction network on the hardmax outputs (one-hot representation), and also tests it on the hardmax outputs of the classification network.
    \item \textbf{S/S} trains the spell-correction network on the softmax outputs, and also tests it on the softmax outputs of the classification network.
    \item \textbf{H/S} trains the spell-correction network on the hardmax outputs, but tests it on the softmax outputs of the classification network.
    \item \textbf{Mix/S} trains the spell-correction network on the mixed hardmax and softmax outputs, and tests it on the softmax outputs of the classification network.
    \item \textbf{\pmb{$\alpha$}S/S} trains the spell-correction network by scaling the logits of the classification network with a random scaling factor $\alpha \in [0, 1000]$ before applying the softmax function. The higher the scaling factor, the more similar the softmax output is to the hardmax output, such that the network learns to cope with hardmax-, as well as softmax outputs and gradations in between those two.
    \item \textbf{S+\pmb{$\epsilon$}/S} trains the spell-correction network with a randomly added amount of noise to the logits of the classification network before applying softmax. We used zero-mean Gaussian noise with variance $0.1$.
\end{itemize}

The results of the workflow without a final statistical spell-correction are summarized in tables \ref{tabular:joint_evaluation_character_acc} and table \ref{tabular:joint_evaluation_word_acc}. As the data set for the German sign language alphabet is short of videos, the purely softmax based S/S approach apparently suffered from overfitting and failed to provide good results. Therefore, as we can see, the different ways to exploit hard- and softmax inputs during training and testing on the softmax inputs only, not only improve the classical H/H approach by $2-4\%$ in terms of character accuracy, and $2-8\%$ in terms of word accuracy but also increases the accuracy compared to the S/S approach. Most importantly, the improvements of H/S, Mix/S, $\alpha$S/S and S$+\epsilon$/S are consistent, i.e., improved the H/H approach in all test scenarios. Interestingly, the word accuracy does not entirely correlate with the character accuracy, but at least close resembles it up to the positive outlier of $\alpha$S/S in TC3.


\bgroup
\def\arraystretch{1.5}
\begin{center}
\begin{table}[!ht]
	\centering
	\begin{tabular}{|l ||c|c|c|c|c|c|}
	\hline
	 & \multicolumn{6}{c|}{\textbf{Character-Accuracy}}\\%
	\textbf{Test case} & H/H & S/S & H/S & Mix/S & $\alpha$S/S & S+$\epsilon$/S \\ \hline \hline%
	TC1 & 77~\%&77~\%& 79~\%& 78~\% & 78~\% & \textbf{80~\%} \\  
	 \hline%
	TC2 & 73~\%&74~\%&75~\%& 76~\%& 74~\%& \textbf{78~\%}\\  
	 \hline
	TC3 & 74~\% & 68~\% &\textbf{78~\%} &74~\%& 76~\%& 76~\% \\  
	\hline%
TC4  & 81~\%& 78~\%& 84~\%& 84~\%& \textbf{85~\%}& \textbf{85~\%}\\%
	 \hline%
	\end{tabular} \newline
	\caption{Joint evaluation of character accuracy of the fingerspelling- and spell-correction network for different types of inputs used for the spell-correction network during training and inference. The proposed exploration of the softmax outputs improves the classical one-hot representations H/H  consistently.}
	\label{tabular:joint_evaluation_character_acc}
\end{table}

\end{center}

\begin{table}[!ht]
    \vspace{5pt}
	\centering
	\begin{tabular}{|l ||c|c|c|c|c|c|}
	\hline
	&  \multicolumn{6}{c|}{\textbf{Word-Accuracy}}\\%
	\textbf{Test case}& H/H & S/S & H/S & Mix/S & $\alpha$S/S & S+$\epsilon$/S\\ \hline \hline%
	TC1  &26~\%&29~\%&33~\% & 34~\%& 31~\%& \textbf{36~\%}\\  
	 \hline
	TC2 & 29~\%& 31~\%&29~\% &30~\%& 29~\%& \textbf{40~\%}
	 \\ \hline
	TC3 & 32~\%& 21~\%& 33~\%&31~\%& \textbf{34~\%}& 33~\% \\   \hline
	TC4 &36~\%& 31~\%&43~\%&\textbf{44~\%}& \textbf{44~\%}& \textbf{44~\%}\\   \hline
	\end{tabular} \newline
	\caption{Joint evaluation of word accuracy of the fingerspelling- and spell-correction network for different types of inputs used for the spell-correction network during training and inference. The proposed exploration of the softmax outputs improves the classical one-hot representations H/H  consistently.}
	\label{tabular:joint_evaluation_word_acc}
\end{table}
\egroup
While the results of the spell-correction network lead to good character accuracy, the word accuracy varies between around 34 to 44 percent. This leads to the assumption that the network mainly produces words which contain one to two incorrect letters. As statistical spell-corrections especially succeed in correcting lightly misspelled words, a second approach with an additional statistical spell-correction that operates on the output of the spell-correction network is tested to increase the word accuracy. The results of these tests are summarized in table \ref{tabular:joint_evaluation_norvig_character_acc} and table \ref{tabular:joint_evaluation_norvig_word_acc}. It can be seen that the character accuracy can be increased by a small factor while the word accuracy almost doubles compared to the approach without an additional statistical spell-correction.

\begin{figure}
    \centering
    \includegraphics[width=0.8\linewidth]{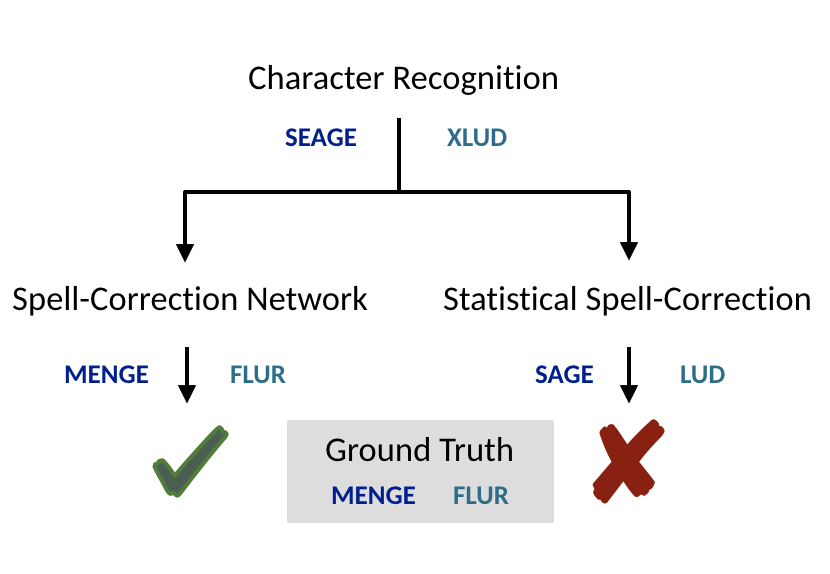}
    \caption{Sample words which show the effectiveness of the proposed approach. While statistical spell-corrections fail in correcting words like `Menge' and `Flur', the output of the spell-correction network matches the ground truth.}
    \label{fig:example_words}
\end{figure}

Comparing the combined spell-correction network and statistical spell-correction approach to the use of the Norvig correction only (show in the `N' column), it is on-par in TC3, but showed significantly worse performance in TC2 and TC4, with the proposed approaches improving the character and word accuracy by up to 20~\% and 25~\%, respectively. 
\bgroup
\def\arraystretch{1.5}
\begin{center}
\begin{table}[!ht]
	\centering
	\begin{tabular}{|c||c|c|c|c|c|c|c|}
	\hline
	& \multicolumn{7}{c|}{\textbf{Char.-Acc. with additional Norvig}}\\%
	\textbf{Test} & H/H & S/S & H/S & Mix/S & $\alpha$S/S & S+$\epsilon$/S & N\\ \hline \hline%
    TC1 & 78~\% &\ 77~\%&\ \textbf{80~\%} &\ 77~\% & 78~\%& \textbf{80~\%} & 78~\%\\ \hline
	TC2 &\ 76~\% &\ 75~\%&\ 76~\% &\ 75~\%& 72~\%& \textbf{80~\%}&\ 61~\%\\ \hline
	TC3 &\ 76~\% &\  72~\%&\ \textbf{77~\%} &\ 73~\%& 73~\%& 73~\%&\ 75~\%\\ \hline%
	TC4 &\ 81~\% &\ 81~\%&\ 85~\% &\ 84~\%& \textbf{89~\%}& 86~\%&\ 77~\%\\ \hline%
	\end{tabular} \newline
	\caption{Joint evaluation of character accuracy of the fingerspelling- and spell-correction network, for different types of inputs used for the spell-correction network during training and inference, in combination with a statistical spell-correction approach of Norvig. The proposed exploration of the softmax outputs improves the classical one-hot representations H/H  consistently.}
	\label{tabular:joint_evaluation_norvig_character_acc}
\end{table}
\end{center}
\begin{table}[!ht]
    \vspace{15pt}
	\centering
	\begin{tabular}{|c||c|c|c|c|c|c|c|}
	\hline
	& \multicolumn{7}{c|}{\textbf{Word-Acc. with Norvig}}\\%
	\textbf{Test} & H/H & S/S & H/S & Mix/S & $\alpha$S/S & S+$\epsilon$/S & N\\ \hline \hline%
	TC1 & 59~\% & 59~\%& 63~\% & 59~\% & 58~\%& \textbf{64~\%} & 63~\%\\ \hline
	TC2 & 58~\% & 56~\%& 59~\% & 56~\%& 51~\%& \textbf{65~\%}& 39~\%\\ \hline
	TC3 & \textbf{61~\%} & 54~\% & 58~\% & 58~\%& 52~\%& 57~\%&\textbf{61~\%}\\ \hline%
	TC4 & 63~\% & 59~\%& 67~\% & 67~\%& \textbf{75~\%}& 70~\%& 61~\% \\ \hline%
	\end{tabular} \newline
	\caption{Joint evaluation of word accuracy of the fingerspelling- and spell-correction network, for different types of inputs used for the spell-correction network during training and inference, in combination with a statistical spell-correction approach of Norvig. The proposed exploration of the softmax outputs improves the classical one-hot representations H/H  consistently.}
	\label{tabular:joint_evaluation_norvig_word_acc}
\end{table}
\egroup
Figure \ref{fig:example_words} illustrates an exemplary qualitative result of the proposed approach. For the inputs "SEAGE" and "XLUD" the statistical spell-correction predicts "SAGE" and "LUD", which are valid German words. The softmax outputs of the classification network ran on the video that resulted in "SEAGE" suggests that while "S" is the most probable first character, "M" is the second most probable one. This additional information allows the spell-correction network to come to a different prediction of the corrected word, which reflects our clues of the input video more closely. 

\begin{figure}
    \centering
    \includegraphics[width=0.8\linewidth]{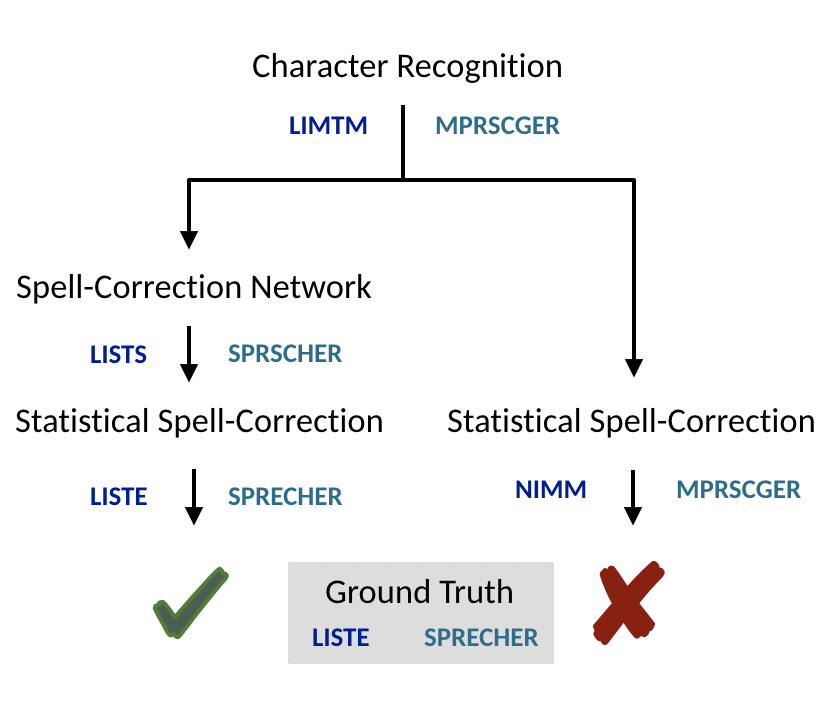}
    \caption{Sample words which show that the combination of spell-correction network and statistical spell-correction result in correctly spelled words, whereas a statistical spell-correction is not able to correct these words.}
    \label{fig:example_words_statistical}
\end{figure}

 Figure \ref{fig:example_words_statistical} furthermore shows that the output of the spell-correction network on the one hand seems to be more human readable than the output of the character recognition, on the other hand, it turns out to be easier correctable for a statistical spell-correction. Turning the word "LIMTM" into "NIMM" or "LISTE" (both of which are valid words) both cause a Levensthein distance of two, but only the additional information of the softmax (e.g. being certain about "L" being the first character and less certain about the last "M") allows to identify the correct word. For a Levenshtein distance $\geq 2$ the statistical spell-correction cannot make any prediction, as seen for "MPRSCGER").

\section{Conclusion}
In this work we have studied how to tailor a spell-correction approach to a specific source of word-prediction, namely a video classifier trained on sign language recognition. We demonstrated that exploiting the softmax probabilities of such a classifier can yield systematic improvements in the spell-correction. Attention has to be paid to the problem of overfitting the spell-corrector to specific examples of softmax probabilities from the classifier, such that mostly training on hardmax outputs but exploiting softmax outputs during inference seems to be favorable - at least in our case of rather scarce training data. By combining learning and statistical methods our final pipeline achieves word accuracies that are up to $25\%$ higher than purely using a statistical approach.

\bibliographystyle{IEEEtran}
\bibliography{refs.bib}

\end{document}